\documentclass[letterpaper, 10 pt, journal, twoside]{IEEEtran}

\IEEEoverridecommandlockouts
\usepackage{kotex}
\usepackage{amssymb}
\usepackage{amsmath}
\usepackage{graphicx}
\usepackage[table,xcdraw]{xcolor}
\usepackage{booktabs}
\usepackage{multirow}
\usepackage{subcaption}
\usepackage{cite}
\usepackage{hyperref}
\usepackage{soul}
\usepackage{comment}
\usepackage{makecell}
\usepackage{booktabs}
\usepackage{multirow}
\usepackage{xcolor}

\usepackage{array}  % \shortstack 사용
\hypersetup{
    colorlinks=true,
    linkcolor=blue,
    filecolor=magenta,
    urlcolor=cyan,
    citecolor=green,
}

\title{\LARGE \bf
SceneAware: Scene-Constrained Pedestrian Trajectory Prediction with LLM-Guided Walkability
}

\author{Juho Bai$^{1}$ and Inwook Shim$^{2,*}$
\thanks{This research was supported by the National Research Foundation of Korea~(NRF) grant funded by the Korea government~(MSIT)~(No.RS-2024-00342176 and RS-2025-02217000.) and INHA UNIVERSITY Research Grant~(2023.) *{Corresponding author: Inwook Shim.}}
\thanks{$^{1}$Juho Bai is with the College of Economics and Business, Hankuk University of Foreign Studies, Republic of Korea
        {\tt\footnotesize juho@hufs.ac.kr}}
\thanks{$^{2}$Inwook Shim is with the Department of Smart Mobility Engineering, Inha University, Republic of Korea {\tt\footnotesize iwshim@inha.ac.kr}}
}

\begin{document}

\maketitle
\thispagestyle{empty}
\pagestyle{empty}

%%%%%%%%%%%%%%%%%%%%%%%%%%%%%%%%%%%%%%%%%%%%%%%%%%%%%%%%%%%%%%%%%%%%%%%%%%%%%%%%
\begin{abstract}
Accurate prediction of pedestrian trajectories is essential for applications in robotics and surveillance systems. While existing approaches primarily focus on social interactions between pedestrians, they often overlook the rich environmental context that significantly shapes human movement patterns. In this paper, we propose SceneAware, a novel framework that explicitly incorporates scene understanding to enhance trajectory prediction accuracy. Our method leverages a Vision Transformer~(ViT) scene encoder to process environmental context from static scene images, while Multi-modal Large Language Models~(MLLMs) generate binary walkability masks that distinguish between accessible and restricted areas during training. We combine a Transformer-based trajectory encoder with the ViT-based scene encoder, capturing both temporal dynamics and spatial constraints. The framework integrates collision penalty mechanisms that discourage predicted trajectories from violating physical boundaries, ensuring physically plausible predictions. SceneAware is implemented in both deterministic and stochastic variants. Comprehensive experiments on the ETH/UCY benchmark datasets show that our approach outperforms state-of-the-art methods, with more than 50\% improvement over previous models. Our analysis based on different trajectory categories shows that the model performs consistently well across various types of pedestrian movement. This highlights the importance of using explicit scene information and shows that our scene-aware approach is both effective and reliable in generating accurate and physically plausible predictions. Code is available at: \url{https://github.com/juho127/SceneAware}
\end{abstract}


%%%%%%%%%%%%%%%%%%%%%%%%%%%%%%%%%%%%%%%%%%%%%%%%%%%%%%%%%%%%%%%%%%%%%%%%%%%%%%%%
\section{INTRODUCTION}

Pedestrian trajectory prediction is a fundamental task in domains such as robotics and surveillance systems aimed at enhancing public safety and services efficiency. Accurate forecasting of human movement allows these systems to operate safely and efficiently in dynamic and crowded environments. Although recent advances have significantly improved the accuracy, many existing methods primarily focus on social interactions among pedestrians, often neglecting the environmental context that critically influences human movement.

In real-world scenarios, pedestrians navigate complex environments shaped by buildings, roads, sidewalks, and other obstacles. These scene elements impose natural constraints on human movement, as people tend to follow walkable paths and avoid physical barriers. Traditional trajectory prediction models that overlook such constraints often generate physically implausible paths, for instance, trajectories that pass through walls or restricted areas.% or barriers.

\begin{figure}[t]
  \centering
  \includegraphics[width=\linewidth]{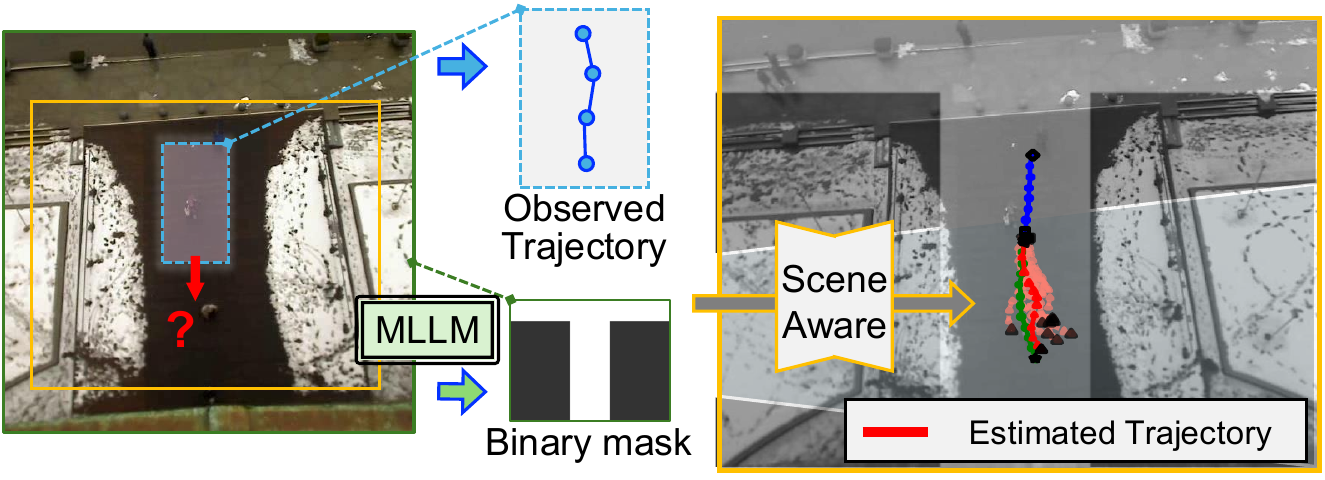}
  \caption{Overview of SceneAware. Inputs: Observed Trajectory and MLLM-generated binary mask that distinguishes walkable from non-walkable areas. Trajectory prediction with Scene Structure Information.~(*MLLM is only for training.)}
  \label{fig:intro_overview}
\end{figure}

Recent advances in deep learning have led to sophisticated trajectory prediction models, from social interaction-aware approaches~\cite{alahi2016social, gupta2018social} to graph-based methods~\cite{mohamed2020social, bae2021disentangled, bae2022npsn}. More recently, language-based approaches such as LMTrajectory~\cite{bae2024lmtrajectory} have emerged, leveraging large language models~(LLMs) to improve motion reasoning.
Despite its impressive performance, they still lack explicit mechanisms for incorporating environmental constraints. As a result, they still continue to produce unrealistic predictions that are misaligned with the actual physical world. Moreover, applying LLMs at every frame incurs substantial computational cost, making them impractical for real-time systems such as autonomous robots or surveillance platforms.

In this paper, we propose SceneAware, a novel framework that explicitly incorporates scene structure understanding into trajectory prediction. Our method employs a pretrained and frozen Vision Transformer~(ViT) scene encoder to distinguish between walkable and non-walkable areas, guiding the model to generate physically plausible trajectories. Figure~\ref{fig:intro_overview} illustrates the overview of our SceneAware framework.
The architecture integrates a Transformer-based trajectory predictor and is guided by a frozen Multi-modal Large Language Model~(MLLM) that generates a walkability-based penalties on the decoder's outputs when predicted trajectories violate inferred physical constraints. Importantly, MLLM is used only during training, allowing the model to learn scene-aware constraints without requiring MLLM inference at test time. This design enables efficient and physically grounded trajectory prediction, making SceneAware suitable for real-time applications such as surveillance and robotics.

We develop both deterministic and stochastic variants of our model, supporting both single- and multi-path forecasting. Experiments on ETH/UCY benchmarks demonstrate that incorporating scene context significantly improves prediction accuracy. Our results show that explicitly scene encoded information provides sufficient spatial understanding without requiring complex pre-processing.
Our contributions can be summarized as follows:
\begin{itemize}
    \item \textbf{Scene-aware trajectory prediction:} We propose a novel pedestrian trajectory prediction framework that jointly encodes pedestrian trajectory and scene structure, enabling physically plausible and context-aware predictions.
    \item \textbf{MLLM-assisted scene encoding:} Our method leverages MLLM to generate a binary walkability mask from a single scene image, eliminating the need for manual annotations or per-frame computation.
    \item \textbf{Alternative evaluation approach:} We introduce a new evaluation that categorizes pedestrian movement patterns for more fine-grained performance analysis.
\end{itemize}

\section{Related Work}

\subsection{Traditional Trajectory Prediction}

Early approaches to trajectory prediction rely on physics-based models~\cite{helbing1995social, pellegrini2009you} that impose pedestrian interactions using human designed force functions. These methods, while interpretable, struggle to capture the complexity of pedestrian behavior in real-world. The Social Force Model~\cite{helbing2000simulating} and its extensions~\cite{zanlungo2011social, hoermann2018dynamic} define the forces of attraction, repulsion, and friction to simulate movement.

With the advent of deep learning, data-driven approaches gained prominence. Recurrent Neural Networks~(RNNs), particularly Long Short-Term Memory~(LSTM) networks, are utilized to the backbone of trajectory prediction models~\cite{alahi2016social,gupta2018social}. Social-LSTM~\cite{alahi2016social} introduces a social pooling mechanism to model interactions between pedestrians, and Social-GAN~\cite{gupta2018social} employs adversarial training to generate more realistic trajectories. Other early deep learning models include those based on vanilla RNNs~\cite{lee2017desire} and encoder-decoder architectures~\cite{sermanet2017attention}. These models primarily focus on learning temporal dependencies in individual trajectories and simple forms of social interaction. However, these early methods also lack robust mechanisms for incorporating scene context, leading to potentially unrealistic predictions.

\subsection{Graph-Based and Attention-Based Models}
Recent years have seen the rise of graph-based approaches~\cite{mohamed2020social,bae2021disentangled,bae2022npsn} that model pedestrian interactions as graphs.
STGAT~\cite{huang2019stgat} and SGCN~\cite{tang2019sr} utilize graph attention and graph convolutional networks, respectively, to model the spatio-temporal relationships between pedestrians. More sophisticated graph-based models consider heterogeneous interactions \cite{shi2020heterogeneous} and future intent \cite{liang2019peeking}.

Attention mechanisms have also been widely adopted in trajectory prediction~\cite{vemula2018social, sadeghian2019sophie, yu2020spatio}. These approaches enable models to focus on relevant parts of the input, such as specific pedestrians or scene elements, improving prediction accuracy. Trajectron++~\cite{salzmann2020trajectron++} uses a graph-based approach with attention mechanism to model interactions and predict multiple future trajectories. Similarly, Social Attention~\cite{vemula2018social} allows each pedestrian to attend to their relevant neighbors. TPNet~\cite{luo2021tpnet} further refines the social attention by incorporating temporal attention. While these graph and attention mechanisms significantly enhance the modeling of complex social dynamics, they often treat the physical environment implicitly and still lack mechanisms to enforce hard constraints.

\begin{figure*}[t]
 \centering
 \includegraphics[width=\linewidth]{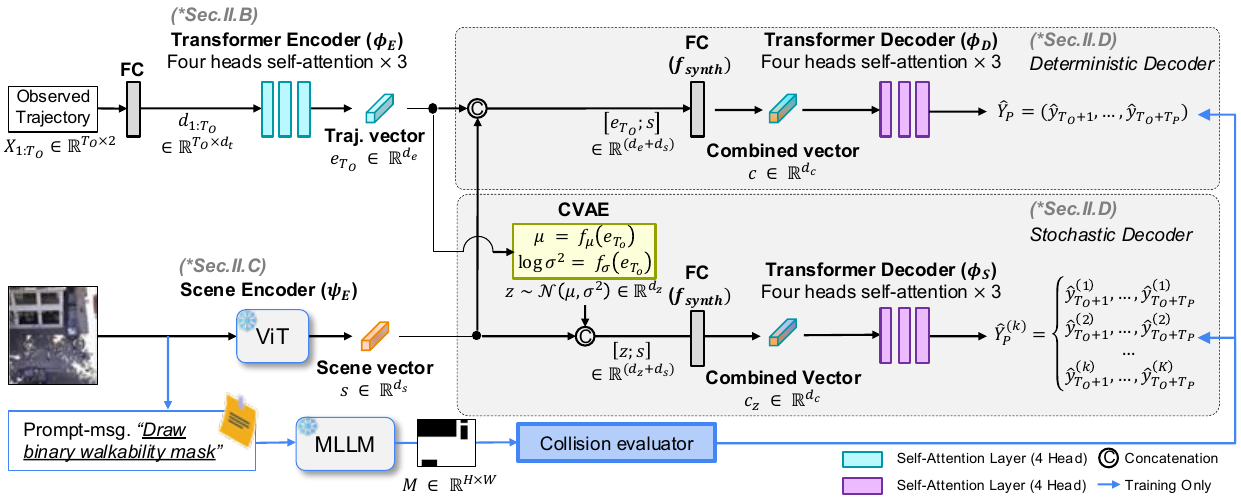}
 \caption{Our SceneAware model separates the encoding of trajectories and scene information. It uses a Transformer to encode the trajectory’s temporal dynamics and a ViT encoder to capture spatial constraints from the scene. These two types of information are then fused before decoding, allowing the model to generate accurate and scene-aware predictions. Binary walkability masks generated by MLLM provide collision constraints for training. This modular design makes it easy to use the best-suited networks for each type of data.}
 \label{fig:architecture}
\end{figure*}

\subsection{Environment-Aware Prediction}

Recognizing the limitations of social-only trajectory prediction, many approaches seek to incorporate environmental context. Initial efforts often do so implicitly, utilizing basic environmental representations like occupancy grids~\cite{trautman2010unfreezing} and rasterized maps~\cite{kuderer2012learning}, learning interaction models such as cost maps via inverse reinforcement learning~\cite{kitani2012activity}. Other approaches continue to integrate context implicitly, employing techniques like scene attention over visual features~\cite{sadeghian2019sophie}, using pre-processed inputs such as semantic segmentation maps~\cite{deo2020trajectory} and updated occupancy grids~\cite{Zhang2023Prescient}.

Most recently, language-based LMTrajectory~\cite{bae2024lmtrajectory} has emerged, recasting trajectory prediction problem as a prompt-based question-answering task, which leverages LLM to understand and predict complex movement patterns, showing impressive performance across benchmark datasets. 

LMTrajectory encodes numerical trajectories into textual prompts, allowing language models to interpret pedestrian movement through natural language understanding. While this leverages the contextual reasoning strengths of large language models, it lacks explicit mechanisms to infer or enforce the geometric and physical constraints of the environment.
%based solely on textified coordinates. 
In contrast, SceneAware addresses this limitation by directly incorporating scene structure constraints, offering a more practical and computationally efficient approach with improved predictive performance.

Moving beyond these implicit methods, research progresses towards explicitly modeling scene constraints and integrating them more directly into the prediction pipeline. One direction involves using conditional generative models~(CVAEs) explicitly conditioned on scene context~\cite{lee2017desire} to ensure physical plausibility. Subsequently, researchers develop methods to more tightly integrate environmental features through spatial refinement modules operating alongside recurrent trajectory encoders~\cite{zhang2019sr}, and convolutional pooling mechanisms designed to jointly process social and scene information~\cite{rhinehart2019precog}. Further advancements include approaches using scene semantics to generate plausible goal locations~\cite{dendorfer2020goal} that condition trajectory generation, and graph-based methods explicitly representing the environment within the graph structure~\cite{chandra2020traphic} interacting with agents. This body of work demonstrates a clear trend towards leveraging explicit scene understanding and constraints.

Our SceneAware builds upon the insights from these works but adopts a distinct strategy centered on task-specific feature learning using a focused representation of the environment. Instead of utilizing richer, but potentially more complex, and computationally intensive representations such as detailed heatmaps~\cite{Ridel2020SceneCompliant}, semantic segmentation maps~\cite{deo2020trajectory}, or depth information~\cite{Qiu2022Egocentric}, SceneAware leverages computationally efficient binary walkability mask. These maps distill the environment down to the essential geometric constraints relevant for navigation.

In addition, several related works have explored many directions, such as improving computational efficiency~\cite{Huang2024Decouple}, refining goal estimation~\cite{Yao2021BiTraP, Zhou2023DACG}, enabling continual~learning~\cite{Wu2023Continual}, and adapting to diverse environments~\cite{Li2024DualAlign}.

\begin{figure*}[t]
 \centering
 \begin{subfigure}{1.0\linewidth}
  \centering
  \includegraphics[width=\linewidth]{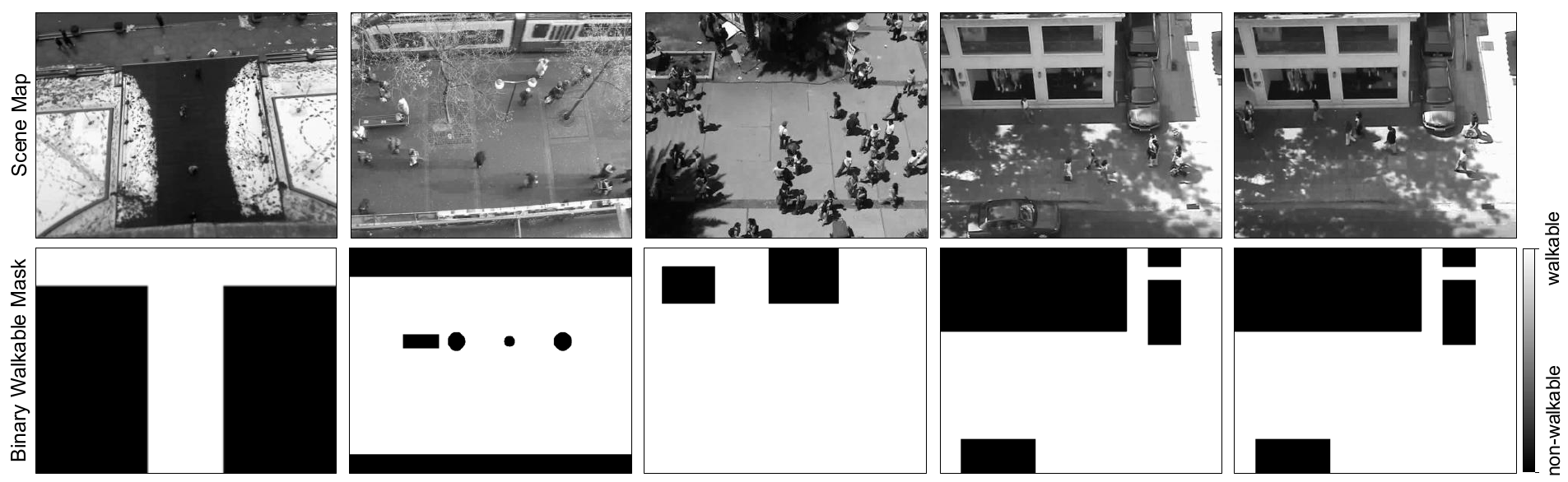}
  \label{fig:binary_mask}
 \end{subfigure} 
 \\
 \vspace{-6mm}
 \begin{subfigure}{0.19\linewidth}
  \centering
  \caption{ETH}
 \end{subfigure}
 \begin{subfigure}{0.19\linewidth}
  \centering
  \caption{HOTEL}
 \end{subfigure}
 \begin{subfigure}{0.19\linewidth}
  \centering
  \caption{UNIV}
 \end{subfigure}
 \begin{subfigure}{0.19\linewidth}
  \centering
  \caption{ZARA1}
 \end{subfigure}
 \begin{subfigure}{0.19\linewidth}
  \centering
  \caption{ZARA2}
 \end{subfigure}
\caption{Results of Scene Map to Binary Walkable Mask conversion across all five datasets. Each pair shows the original top-down view~(top) and the corresponding generated binary walkable map~(bottom) by MLLM. White areas in the binary masks represent walkable regions, while black indicates non-walkable areas. Note how the binary masks simplify complex visual information into clear environmental constraints, focusing only on regions relevant for pedestrian navigation.}
 \label{fig:mask_examples}
\end{figure*}

\section{Methodology}

\subsection{Problem Formulation}

Given a sequence of observed positions $X_{1:T_O} = \{x_1, x_2, ..., x_{T_{O}}\}$ for a pedestrian, where $x_t \in \mathbb{R}^2$ represents the 2D coordinates at time step $T_O$, the trajectory prediction task aims to forecast the future positions $Y_{1:T_P} = \{y_1, y_2, ..., y_{T_{P}}\}$, where $y_t \in \mathbb{R}^2$ are the coordinates at future time steps $T_P$. Here, the pedestrian's position is expressed in the input image coordinate system.

Our objective is to generate future trajectories that align with observed motion patterns while considering the physical scene structure constraints of the environment.

The model architecture, illustrated in Fig.~\ref{fig:architecture}, is intentionally designed with three core components to effectively integrate observed trajectories and environmental context: 1)~Trajectory Encoder: Captures how the pedestrian has been moving. 2)~Scene Encoder: Understands the physical constraints of the environment. 3)~Trajectory Decoder: Combines motion history and scene context to generate future steps. This separation allows each component to specialize. 
Specifically, we guide the LLM to convert the input image into a binary scene mask that clearly indicates walkability regions. This helps the decoder generate future trajectories based on clearer and more explicit environmental information.

\subsection{Trajectory Encoder}

Our model utilizes the Transformer-based encoder \cite{yuan2021agentformer,Huang2024Decouple}, leveraging its advantages for modeling complex temporal dependencies and long-range interactions in the sequence of observed positions. The trajectory encoder $\Phi_E$ processes the raw trajectory coordinates by first transforming each observed 2D coordinate $x_t$ into a higher-dimensional embedding vector $d_t$ using a fully connected layer with non-linear activation. This embedding step allows the model to learn a richer representation of spatial positions compared to using raw coordinates directly. The complete embedded trajectory sequence $d_{1:T_O}$ is then processed by the Transformer encoder to produce the final encoded representation:
\begin{equation}
e_{T_O} = \Phi_E(d_{1:T_{O}}),
\end{equation}
where $e_{T_O}\in\mathbb{R}^{d_e}$ indicates the encoded observed trajectory feature. The Transformer's self-attention mechanism weighs the importance of different positions, focusing dynamically on the most relevant parts for prediction. This embedding transforms coordinates into a high-dimensional representation that generalizes well across diverse environments without overfitting. In our encoder-decoder architecture, $e_{T_O}$ is combined with scene structure information before being fed into the decoder, providing integrated context for accurate and scene-aware trajectory prediction.

\subsection{Scene Encoder}
\label{sec:scene_encoder}

To encode the scene structure information, our framework employs a pretrained ViT as the scene encoder to generate the feature vector~$s\in\mathbb{R}^{d_s}$. The ViT captures spatially global feature relationships across the entire image, allowing the decoder to reason effectively about how the observed trajectories related to important environmental features such as walkable corridors, obstacles, and boundaries. The dimension of the scene embedding, $d_s$, is set to match the trajectory embedding dimension $d_e$, enabling straightforward fusion with the trajectory features in the decoder.

In addition to the ViT-based scene encoding, our SceneAware framework includes a penalty mechanism that discourages implausible trajectory predictions by referencing a binary walkability mask. This penalty is applied alongside the primary loss fuction, as described in Sec.~\ref{sec:loss}. We design prompt-based queries such as \textit{``Generate a binary walkability mask from this scene image, with white for walkable areas and black for obstacles.''} that guide the MLLM~\cite{theC3} to distinguish walkable from non-walkable regions. This approach enables the model to learn scene structure representations without human supervision, while still enforcing physical constraints essential for realistic path prediction. Examples of the generated binary walkability masks are shown in Fig.~\ref{fig:mask_examples}.

\begin{table*}[t]
\centering
\caption{Performance comparisons for deterministic and stochastic models. All values are in meters. The symbol `-' indicates that the performance evaluation is not reported. The best performance is highlighted in \textbf{bold}, and \underline{underline} indicates the best performance among the baseline methods, excluding our method.}
\label{tab:main_comparison}
\resizebox{1.0\textwidth}{!}
{
\begin{tabular}{lcccccc|cccccc}
\toprule
& \multicolumn{6}{c}{\textbf{Deterministic Models~(ADE/FDE)}} & \multicolumn{6}{c}{\textbf{Stochastic Models~(minADE$_{20}$/minFDE$_{20}$)}} \\
\midrule
\textbf{Method} & \textbf{ETH} & \textbf{HOTEL} & \textbf{UNIV} & \textbf{ZARA1} & \textbf{ZARA2} & \textbf{AVG}  & \textbf{ETH} & \textbf{HOTEL} & \textbf{UNIV} & \textbf{ZARA1} & \textbf{ZARA2} & \textbf{AVG} \\
\midrule
Social-LSTM~\cite{alahi2016social} & 1.09/2.35 & 0.79/1.76 & 0.67/1.40 & 0.47/1.00 & 0.56/1.17 & 0.72/1.54 & - & - & - & - & - & - \\
Social-GAN~\cite{gupta2018social} & 0.81/1.52 & 0.72/1.61 & 0.60/1.26 & 0.34/0.69 & 0.42/0.84 & 0.58/1.18 & 0.58/1.18 & 0.39/0.66 & 0.44/0.76 & 0.31/0.60 & 0.29/0.53 & 0.40/0.74 \\
Social-STGCNN~\cite{mohamed2020social} & 0.64/1.11 & 0.49/0.85 & 0.44/0.79 & 0.34/0.53 & 0.30/0.48 & 0.44/0.75 & - & - & - & - & - & - \\
DMRGCN~\cite{bae2021disentangled} & 0.59/1.04 & 0.24/0.41 & 0.39/0.74 & 0.29/0.54 & 0.28/0.52 & 0.36/0.65 & - & - & - & - & - & - \\
NPSN~\cite{bae2022npsn} & 0.56/0.97 & 0.21/0.37 & 0.36/0.71 & 0.28/0.53 & 0.26/0.49 & 0.33/0.61 & - & - & - & - & - & - \\
STGAT~\cite{huang2019stgat} & - & - & - & - & - & - & 0.39/0.66 & 0.29/0.47 & 0.35/0.56 & 0.29/0.52 & 0.25/0.45 & 0.30/0.53 \\
BiTraP-NP~\cite{Yao2021BiTraP} & - & - & - & - & - & - & 0.38/0.82 & 0.11/0.22 & 0.21/0.41 & 0.14/0.27 & 0.11/0.22 & 0.18/0.35 \\
DACG~\cite{Zhou2023DACG} & - & - & - & - & - & - & 0.35/0.65 & 0.12/0.23 & 0.17/0.36 & 0.17/0.32 & 0.13/0.25 & 0.19/0.38 \\
DGCN+STDec~\cite{Bhujel2023Disentangle} & 0.54/\underline{0.88} & 0.24/0.36 & \underline{0.27}/\underline{0.48} & \underline{0.21}/\underline{0.42} & \underline{0.20}/0.38 & \underline{0.29}/\underline{0.50} & \underline{0.31}/\underline{0.54} & 0.12/0.21 & \underline{0.13}/\underline{0.26} & \underline{0.10}/\underline{0.19} & \underline{0.10}/\underline{0.19} & \underline{0.15}/\underline{0.28} \\
GA-STT~\cite{Zhou2022GASTT} & \underline{0.51}/0.89 & 0.22/0.38 & 0.33/0.66 & 0.24/0.46 & \underline{0.20}/\underline{0.33} & 0.30/0.54 & - & - & - & - & - & - \\
LMTrajectory~\cite{bae2024lmtrajectory} & 0.71/1.30 & \underline{0.10}/\underline{0.18} & 0.33/0.67 & 0.44/0.94 & 0.80/1.31 & 0.48/0.88 & 0.36/0.56 & \underline{0.08}/\underline{0.12} & 0.18/0.32 & 0.28/0.44 & 0.18/0.25 & 0.22/0.32 \\
\midrule
SceneAware~(raw) & 0.06/0.10 & 0.13/0.23 & \textbf{0.06}/\textbf{0.08} & \textbf{0.03}/\textbf{0.06} & \textbf{0.07}/\textbf{0.14} & 0.070/0.121 & 0.11/0.22 & \textbf{0.06}/\textbf{0.08} & \textbf{0.04}/\textbf{0.07} & 0.09/0.14 & \textbf{0.05}/\textbf{0.08} & 0.068/0.119 \\
SceneAware~(mask) & \textbf{0.05}/\textbf{0.08} & \textbf{0.03}/\textbf{0.05} & \textbf{0.06}/0.11 & 0.06/0.10 & 0.11/0.20 & \textbf{0.063}/\textbf{0.109} & \textbf{0.09}/\textbf{0.16} & 0.07/0.13 & 0.05/0.10 & \textbf{0.04}/\textbf{0.06} & 0.06/0.10 & \textbf{0.061}/\textbf{0.110} \\
\bottomrule
\end{tabular}
}
\end{table*}

\subsection{Trajectory Decoder}
The decoder's primary role is to generate the future trajectory sequence $Y$ utilizing two input sources: the encoded motion patterns in the trajectory context vector~$e_{T_O}$ from the Trajectory Encoder, and the encoded scene structure in the scene feature vector~$s$ from the Scene Encoder. Trajectory context~$e_{T_O}$ is concatenated with scene feature vector~$s$. A subsequent linear transformation yields a unified context embedding that conditions the decoder for generating spatially consistent predictions. Based on prior approaches~\cite{gupta2018social, sadeghian2019sophie}, our model includes both deterministic and stochastic decoders.

\noindent\textbf{Deterministic Model} integrates trajectory context $e_{T_O}$ and scene context $s$ to predict the most likely future path sequence. First, these contexts are concatenated, denoted as $[e_{T_O}; s]$. A linear fusion layer $f_{synth}$ then processes this combined vector to produce a unified context embedding $c = f_{synth}([e_{T_O}; s])$. The decoder transformer $\Phi_D$ takes this embedding to predict the future path sequence $\hat{Y}_{P} = \Phi_D(c)$, where $\hat{Y}_{P} = (\hat{y}_{T_{O+1}}, ..., \hat{y}_{T_{O}+T_{P}})$ with $\hat{y}_{t} \in \mathbb{R}^2$ representing the predicted position at each future timestep $t$.

\noindent\textbf{Stochastic Model} adopts the conditional variational autoencoder~(CVAE) approach~\cite{salzmann2020trajectron++} to estimate multiple plausible future paths. CVAE uses the trajectory context vector~$e_{T_O}$ to parameterize a Gaussian distribution~$\mathcal{N}(\mu, \sigma^2)$ over a latent variable $z \in \mathbb{R}^{d_z}$, where $\mu = f_{\mu}(e_{T_{O}})$ and $\log\sigma^2 = f_{\sigma}(e_{T_{O}})$. $f_{\mu}$ and $f_{\sigma}$ are fully connected layers mapping $e_{T_{O}}$ to the mean and log-variance of the latent distribution. To generate a prediction sample, we first sample a latent variable: $z \sim \mathcal{N}(\mu, \sigma^2)$. This latent variable is then concatenated with $s$, denoted as $[z; s]$. A dedicated linear synthesis layer $f_{synth}$ synthesizes the decoder conditioning embedding $c_z = f_{synth}([z; s])$ from this combined vector. Finally, the decoder transfomer $\Phi_S$ generates a future path sequence sample $\hat{Y}_{P}^{(k)} = \Phi_S(c_z)$. Multiple samples can be generated by repeating the sampling and decoding steps.

\subsection{Collision Penalty and Loss Function}
\label{sec:loss}

Our training objective is designed to enable end-to-end learning of both trajectory prediction and scene structure understanding. To enforce adherence to the physical constraints of the environment, we introduce a collision penalty that discourages predicted trajectories from intersecting with non-walkable regions. Given the binary walkability mask $M\in\{0,1\}^{H\times W}$, the collision penalty is defined as:
\begin{equation}
\mathcal{L}_{\text{C}} = \lambda_{\text{C}} \sum_{t=T_O+1}^{T_O+T_{\text{P}}} \mathcal{C}(\hat{y}_t, M),
\label{eq:collision_penalty}
\end{equation}
where $\mathcal{C}(\hat{y}_t, M) = 1$ if the predicted position $\hat{y}_t$ falls within non-walkable areas ($M < 0.5$) or outside image bounds, and $0$ otherwise. The hyperparameter $\lambda_{\text{C}}$ controls the trade-off between trajectory accuracy and collision avoidance.

\noindent\textbf{Deterministic Model.} We use the standard mean squared error~(MSE) loss between the predicted absolute positions $\hat{y}_t$ and the ground truth absolute positions $y_t$ over the prediction horizon as used in method~\cite{Wu2023Continual}:
\begin{equation}
\mathcal{L}_{\text{D}} = \frac{1}{T_{P}} \sum_{t=T_{O}+1}^{T_{O}+T_{P}} || \hat{y}_t - y_t ||^2.
\label{eq:loss_det}
\end{equation}
This loss directly penalizes the Euclidean distance between the prediction and the ground truth at each future step, encouraging the model to produce a single trajectory that closely matches the actual future path. 

The final objective function is defined as the sum of the deterministic loss and the collision penalty term.
\begin{equation}
\mathcal{L}_{\text{D+C}} = \mathcal{L}_{\text{D}} + \mathcal{L}_{\text{C}}
\label{eq:loss_det_collision}
\end{equation}

\noindent\textbf{Stochastic Model.} The loss function needs to achieve two goals: ensuring the predictions are accurate and encouraging diversity among the generated samples, while also regularizing the latent space. We adopt the compound losses used in~\cite{Zhou2023DACG}:
\begin{equation}
\mathcal{L}_{\text{S}} = \mathcal{L}_{\text{best}} + \lambda_{KL} \mathcal{L}_{KL},
\label{eq:loss_stoch}
\end{equation}
where $\mathcal{L}_{\text{best}}$ is the best-of-$K$ loss, defined as the minimum MSE over the $K$ generated samples:
\begin{equation}
\mathcal{L}_{\text{best}} = \min_{k \in \{1,...,K\}} \frac{1}{T_{P}} \sum_{t=T_{O}+1}^{T_{O}+T_{P}} || \hat{y}_t^k - y_t ||^2.
\label{eq:loss_best}
\end{equation}
This loss encourages the model to produce at least one sample trajectory~$\hat{y}^k$ that is close to the ground truth $y_t$, effectively promoting diversity. It acknowledges that predicting the single exact future is hard, but covering the ground truth with one of the samples is achievable.
$\mathcal{L}_{KL}$ is the Kullback-Leibler~(KL) divergence between the learned latent distribution $\mathcal{N}(\mu, \sigma^2)$ and a prior distribution, typically the standard normal distribution $\mathcal{N}(0, I)$:
\begin{equation}
\mathcal{L}_{KL} = D_{KL}(\mathcal{N}(\mu, \sigma^2) || \mathcal{N}(0, I)).
\label{eq:loss_kl}
\end{equation}
This KL divergence term acts as a regularizer, encouraging the learned latent distributions to stay close to the prior. This helps prevent the posterior collapse, a situation in which the standard deviation $\sigma$ becomes zero, and ensures the latent space remains well-structured to support meaningful sampling. The hyperparameter $\lambda_{KL}$ controls the trade-off between the prediction accuracy and diversity, represented by the loss term~$\mathcal{L}_{\text{best}}$, and latent space regularization~$\mathcal{L}_{KL}$.

The final loss combines the collision penalty function with the stochastic prediction loss.
\begin{equation}
\mathcal{L}_{\text{S+C}} = \mathcal{L}_{\text{best}} + \lambda_{KL} \mathcal{L}_{KL} + \mathcal{L}_{\text{C}}
\label{eq:loss_stoch_collision}
\end{equation}

\begin{table*}[tb]
\centering
\caption{Trajectory category-wise performance comparison between deterministic and stochastic models. Values represent ADE performance in meters using weighted average across all trajectory samples. \textbf{Bold values} indicate the best performance improvement for each category.}
\label{tab:trajectory_category_performance}
\resizebox{1.0\textwidth}{!}{
\begin{tabular}{ll|r|r|r|r|r|r|r|r|r|r|r|r}
\toprule
& & \multicolumn{6}{c|}{\textbf{Deterministic Models (ADE)}} & \multicolumn{6}{c}{\textbf{Stochastic Models (ADE)}} \\
\midrule
\textbf{Method} & \textbf{Category} & \textbf{ETH} & \textbf{HOTEL} & \textbf{UNIV} & \textbf{ZARA1} & \textbf{ZARA2} & \textbf{Avg} & \textbf{ETH} & \textbf{HOTEL} & \textbf{UNIV} & \textbf{ZARA1} & \textbf{ZARA2} & \textbf{Avg} \\
\midrule
\multirow{4}{*}{\shortstack{Social-GAN\\~\cite{gupta2018social}}} 
& Straight & 0.897 & 0.221 & 0.533 & 0.793 & 0.604 & 0.610 & 0.865 & 0.177 & 0.532 & 0.754 & 0.562 & 0.578 \\
& Turning & 0.900 & 0.223 & 0.517 & 0.792 & 0.602 & 0.607 & 0.878 & 0.174 & 0.517 & 0.758 & 0.562 & 0.578 \\
& HighVar & 0.897 & 0.200 & 0.526 & 0.799 & 0.622 & 0.609 & 0.877 & 0.184 & 0.526 & 0.785 & 0.558 & 0.586 \\
& Circling & 0.895 & 0.215 & 0.519 & 0.777 & 0.605 & 0.602 & 0.859 & 0.182 & 0.519 & 0.767 & 0.561 & 0.578 \\
\midrule
\multirow{4}{*}{\shortstack{DGCN+STDec\\~\cite{Bhujel2023Disentangle}}} 
& Straight & 0.559 & 0.239 & 0.272 & 0.213 & 0.201 & 0.297 & 0.321 & 0.120 & 0.131 & 0.101 & 0.100 & 0.155 \\
& Turning & 0.529 & 0.242 & 0.268 & 0.205 & 0.198 & 0.288 & 0.304 & 0.121 & 0.129 & 0.098 & 0.099 & 0.150 \\
& HighVar & 0.471 & 0.239 & 0.272 & 0.226 & 0.201 & 0.282 & 0.270 & 0.119 & 0.131 & 0.108 & 0.100 & 0.146 \\
& Circling & 0.632 & 0.239 & 0.267 & 0.236 & 0.201 & 0.315 & 0.363 & 0.120 & 0.128 & 0.112 & 0.101 & 0.165 \\
\midrule
\multirow{4}{*}{\shortstack{SceneAware\\(raw)}}
& Straight & 0.042 & 0.131 & 0.060 & 0.025 & 0.073 & 0.066 & 0.097 & 0.057 & 0.034 & 0.089 & 0.047 & 0.065 \\
& Turning & 0.217 & 0.122 & 0.064 & 0.057 & 0.091 & 0.110 & 0.208 & 0.074 & 0.047 & 0.089 & 0.063 & \textbf{0.096} \\
& HighVar & 0.054 & 0.112 & 0.060 & 0.035 & 0.074 & 0.067 & 0.114 & 0.057 & 0.036 & 0.081 & 0.047 & 0.067 \\
& Circling & 0.028 & 0.113 & 0.061 & 0.025 & 0.075 & 0.060 & 0.091 & 0.056 & 0.034 & 0.073 & 0.044 & 0.060 \\
\midrule
\multirow{4}{*}{\shortstack{SceneAware\\(mask)}}
& Straight & 0.034 & 0.025 & 0.065 & 0.063 & 0.107 &\textbf{0.059} & 0.079 & 0.071 & 0.044 & 0.034 & 0.055 & \textbf{0.057} \\
& Turning & 0.213 & 0.062 & 0.066 & 0.058 & 0.118 & \textbf{0.103} & 0.226 & 0.083 & 0.056 & 0.054 & 0.070 & 0.098 \\
& HighVar & 0.077 & 0.029 & 0.062 & 0.047 & 0.105 & \textbf{0.064} & 0.072 & 0.071 & 0.049 & 0.031 & 0.055 & \textbf{0.056} \\
& Circling & 0.050 & 0.027 & 0.064 & 0.028 & 0.107 & \textbf{0.055} & 0.065 & 0.070 & 0.043 & 0.025 & 0.052 & \textbf{0.051} \\
\bottomrule
\end{tabular}
}
\end{table*}

\section{Experiments}

In this section, we conduct quantitative evaluations to demonstrate the effectiveness of our SceneAware approach, examining both deterministic and stochastic models. We also compare our performance with state-of-the-art methods.

\subsection{Implementation Details}
\label{sec:imp_details}
We implement SceneAware network using Pytorch framework~\cite{paszke2019pytorch}, utilizing Adam optimizer~\cite{kingma2014adam} with a learning rate of $0.001$. For a fair performance comparison with other state-of-the-arts, the observation time~$T_O$ is $8$~frames~($3.2s$) and prediction time~$T_{P}$ is $12$~frames~($4.8s$) following standard practice in the field~\cite{alahi2016social, gupta2018social}. The collision penalty weight $\lambda_{\text{C}}$ is set to 30.0 across all experiments. For the stochastic model, the KL divergence weight~$\lambda_{KL}$ and the number of generated samples $K$ are set to $0.1$ and $20$, respectively. The details of our SceneAware network is illustrated in Fig.~\ref{fig:architecture} and github repo~\url{https://github.com/juho127/SceneAware}

For evaluation, we utilize the standard benchmark datasets widely used in pedestrian trajectory prediction: ETH~\cite{pellegrini2009you} and UCY~\cite{lerner2007crowds}, which contain pedestrian movements recorded in diverse real-world environments. Following common practice in prior state-of-the-art methods, we adopt \textit{K}-fold validation~(\textit{K}=5) to ensure fair evaluation.

\subsection{Quantitative Evaluation}

In this evaluation, we use common quantitative measures of trajectory prediction: Average Displacement Error~(ADE) and Final Displacement Error~(FDE).
Table \ref{tab:main_comparison} presents a comprehensive performance comparison between our proposed SceneAware model and existing state-of-the-art methods across both deterministic and stochastic prediction. The compared algorithms span multiple architectural approaches, categorized based on their fundamental design principles.

\begin{figure}[t]
  \centering
  \includegraphics[height=4.15cm, width=1.0\linewidth]{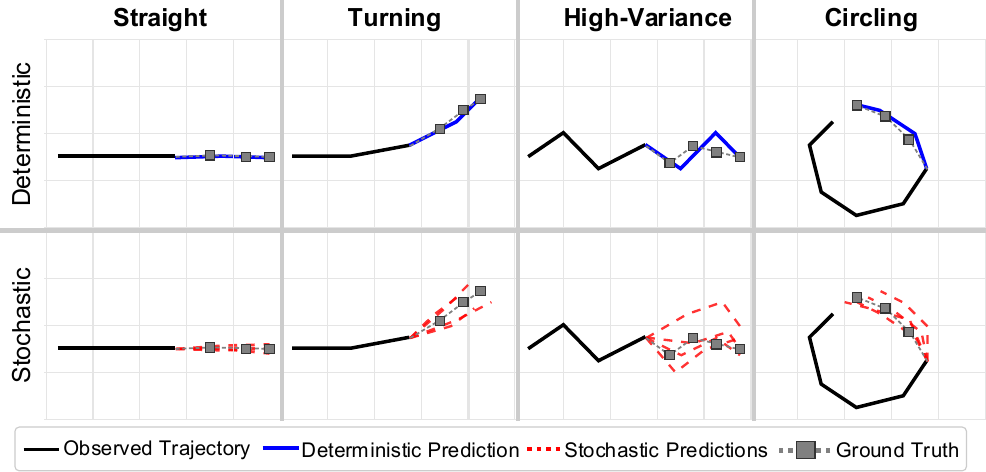}
  \caption{
  Examples of pedestrian trajectory categorization according to distinct pedestrian movement patterns.
  }
  \label{fig:traj_categorization}
\end{figure}

\subsubsection{RNN/LSTM-based Social Models}
Early approaches like Social-LSTM~\cite{alahi2016social} and Social-GAN~\cite{gupta2018social} leverage recurrent architectures with specialized social pooling mechanisms to model pedestrian interactions. These models lack explicit environmental understanding in their structure.
Our SceneAware model reduces error by approximately $85\%$ compared to Social-LSTM, highlighting the critical importance of incorporating scene context into prediction models.

\subsubsection{Graph-based Models}
The second category comprises graph-based approaches including Social-STGCNN~\cite{mohamed2020social}, DMRGCN~\cite{bae2021disentangled}, NPSN~\cite{bae2022npsn}, and STGAT~\cite{huang2019stgat}, which model pedestrians and their interactions using graph structures. While effective at capturing interpersonal dynamics, 
These graph models effectively capture interpersonal relationships but still lack crucial environmental information needed for accurate trajectory prediction. SceneAware's explicit scene modeling provides approximately $67\%$ error reduction compared to the best graph-based model~(NPSN), demonstrating the advantage of clear environmental representation.

\begin{figure}[t]
    \centering
    \includegraphics[height=6cm, width=1.0\linewidth, keepaspectratio]{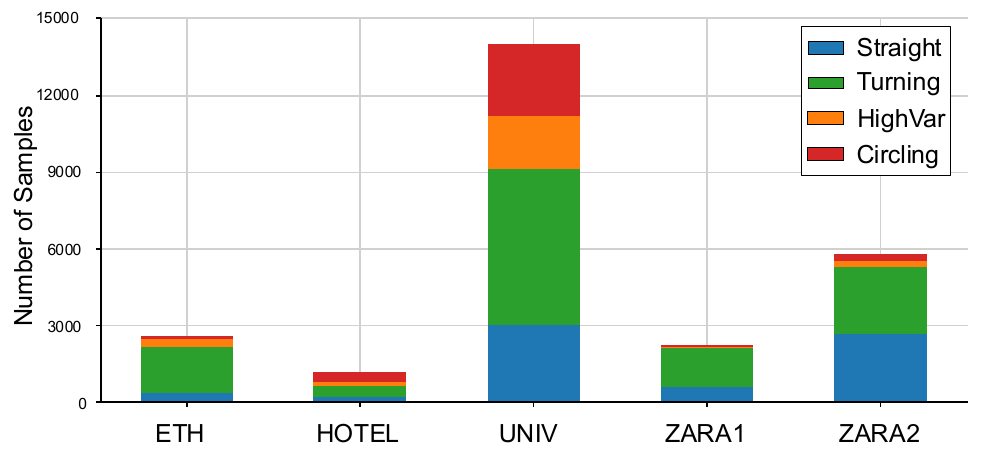}
    \caption{The number of samples and the distribution of trajectory categories across the benchmark datasets. }
    \label{fig:sample_distribution}
\end{figure}

\begin{figure*}[t]
 \centering
 \begin{subfigure}{1.0\linewidth}
  \centering
  \includegraphics[width=\linewidth]{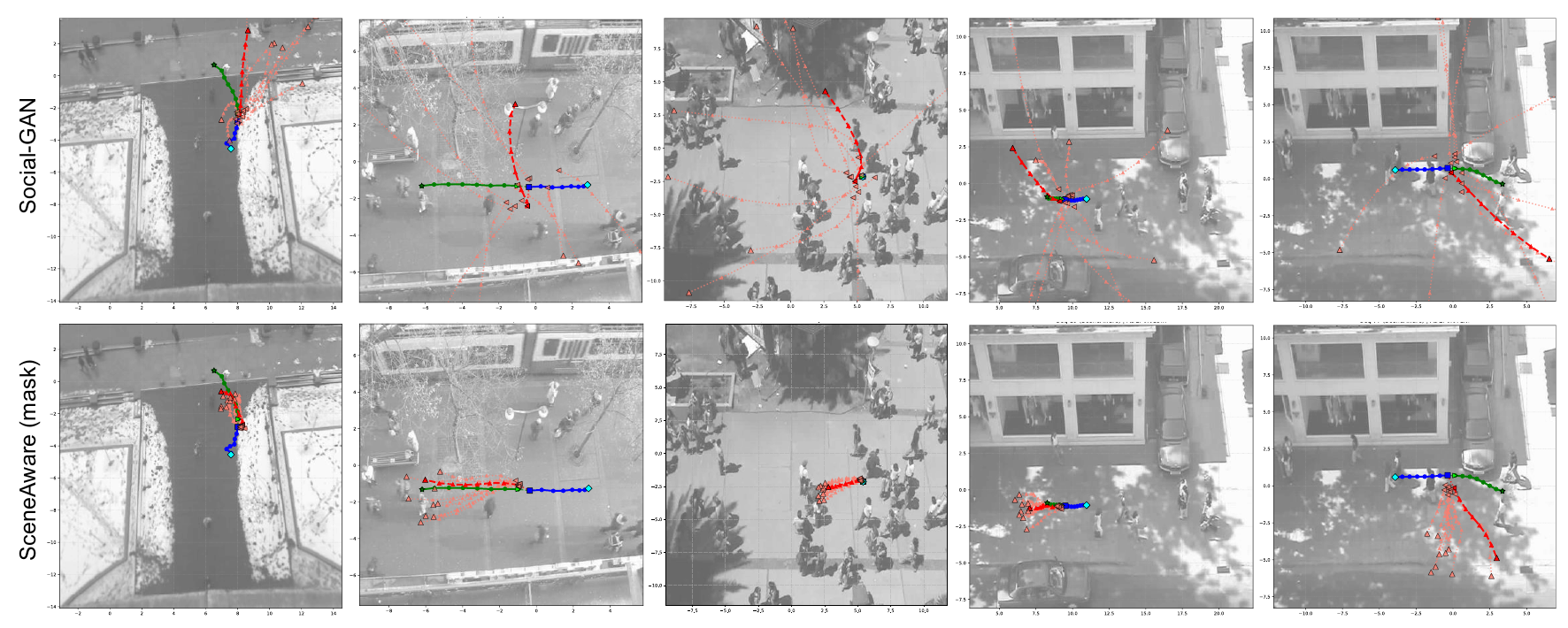}

 \end{subfigure} 
 \\
 \vspace{-1.0mm}
 \begin{subfigure}{0.19\linewidth}
  \centering
  \caption{ETH}
 \end{subfigure}
 \begin{subfigure}{0.19\linewidth}
  \centering
  \caption{HOTEL}
 \end{subfigure}
 \begin{subfigure}{0.19\linewidth}
  \centering
  \caption{UNIV}
 \end{subfigure}
 \begin{subfigure}{0.19\linewidth}
  \centering
  \caption{ZARA1}
 \end{subfigure}
 \begin{subfigure}{0.19\linewidth}
  \centering
 \caption{ZARA2}

 \end{subfigure}
\caption{Qualitative comparisons of SceneAware and Social-GAN stochastic model across all trajectory categories: Blue line~(observed trajectory), Green line~(ground truth future) and Red lines~(predicted samples, the best one is bold.)}
 \label{fig:qualitative}
\end{figure*}

\subsubsection{Goal and Scene-oriented Models}
More recent approaches incorporate goal estimation and some degree of scene understanding, such as BiTraP-NP~\cite{Yao2021BiTraP}, DACG~\cite{Zhou2023DACG}, DGCN+STDec~\cite{Bhujel2023Disentangle}, and GA-STT~\cite{Zhou2022GASTT}. 
DGCN+STDec semantically disentangles graph information into temporal factors~(\textit{e.g.},~velocity) and spatial factors~(\textit{e.g.},~interpersonal positioning). They leverage interpersonal information to implicitly reflect environmental constraints.
Our SceneAware model improves upon DGCN+STDec by approximately $58\%$ for deterministic predictions, demonstrating that our direct binary walkability representation provides more effective scene constraints than the implicit or partial scene understanding in these approaches.

\subsubsection{Language-based Models}
LMTrajectory~\cite{bae2024lmtrajectory} represents an innovative approach using large language models to interpret trajectory patterns. Despite leveraging powerful contextual understanding, this approach struggles to fully capture geometric constraints from textified coordinates alone. SceneAware provides approximately $77\%$ improvement over LMTrajectory in deterministic prediction, suggesting that explicit geometric representation of walkable areas offers advantages that even sophisticated language models cannot match through text-based coordinates.

\subsubsection{Scene Representation Analysis}
Our quantitative evaluation also reveals important insights about scene structure constrained approaches. In the Table.~\ref{tab:main_comparison}, SceneAware~(raw) utilizes scene information without generating the binary walkable mask map. SceneAware~(mask) tends to perform better the SceneAware~(raw) with a $31\%$ reduction in the average error. This shows that simplified binary walkability masks provide clearer, more explicit environmental constraints than raw scene images with potentially distracting visual details.

Overall, our SceneAware model consistently outperforms all previous approaches across most benchmark datasets, demonstrating the critical importance of explicit scene understanding in trajectory prediction systems.

\subsection{Analysis of Trajectory Categorization}
\label{sec:anal_category}

We analyze how the performance of the model varies in different patterns of pedestrian movement. As illustrated in Fig.~\ref{fig:traj_categorization}, the benchmark datasets include four distinct trajectory categories: Straight, Turning, High-Variance, and Circling. The number of samples per category is different within each dataset~(see~Fig.~\ref{fig:sample_distribution}.) Such imbalanced distributions make it difficult to evaluate performance trends solely based on overall metrics. In fact, Table~\ref{tab:main_comparison} does not allow category-wise performance analysis, highlighting the need for a more fine-grained evaluation.
To examine the performance across these categories, we compare our SceneAware model with the baseline DGCN+STDec~\cite{Bhujel2023Disentangle} and Social-GAN~\cite{gupta2018social} in Table~\ref{tab:trajectory_category_performance}.

The results demonstrate that SceneAware achieves substantial improvements across all trajectory categories for both deterministic and stochastic prediction. SceneAware~(mask) consistently outperforms the previous works across all datasets and categories.
Notably, our SceneAware model maintains stable performance across all categories, which indicates that its encoded scene understanding effectively captures diverse pedestrian behaviors and supports robust prediction across varying levels of trajectory complexity.

\subsection{Qualitative Evaluation}

We select Social-GAN~\cite{gupta2018social} for qualitative comparison, as it represents the best performing model among publicly available algorithms that do not explicitly use scene information, providing an ideal baseline to demonstrate the effectiveness of our scene structure constraint.
Figure~\ref{fig:qualitative} shows clear differences between the stochastic predictions of SceneAware and Social-GAN. 
Through our explicit environmental constraint learning, SceneAware's stochastic distributions converge more directionally. In structured environments like ETH, HOTEL, and ZARA while Social-GAN's predicted samples~(red dashed lines) exhibit wide dispersions that violate physical constraints, SceneAware maintains appropriate uncertainty while respecting environmental boundaries.

\section{Conclusion}
We introduced SceneAware, a novel framework that integrates image-based scene understanding into pedestrian trajectory prediction. By combining temporal motion dynamics from past trajectories with spatial constraints derived from static scene images, SceneAware enhances prediction accuracy through a simple yet effective fusion strategy.
Extensive experiments on the ETH/UCY benchmark datasets demonstrate that SceneAware significantly improves performance over existing state-of-the-art methods. These results highlight the importance of incorporating scene context for generating accurate and physically plausible trajectories. The effectiveness of SceneAware, despite its simplicity, suggests strong potential for real-world applications in robotics, autonomous systems, and intelligent surveillance.

A promising direction for future work is to develop a hybrid framework that combines scene-level information with the trajectory history of each pedestrian as input to LLM. By integrating these two complementary sources, explicit geometric constraints from the scene and rich temporal context from motion history, the system could achieve a deeper semantic understanding of pedestrian behavior. We also envision integrating online adaptation capabilities to support deployment in continuously evolving environments. Overall, SceneAware offers a strong foundation for future research in scene-aware human trajectory forecasting.

\bibliographystyle{IEEEtran}
\bibliography{references}

\end{document}